\documentclass[conference]{IEEEtran}
\IEEEoverridecommandlockouts
\usepackage{cite}
\usepackage{amsmath,amssymb,amsfonts}
\usepackage{algorithmic}
\usepackage{graphicx}
\usepackage{textcomp}
\usepackage{xcolor}
\usepackage{hyperref}
\usepackage{textcomp}
\usepackage{url}

\def\BibTeX{{\rm B\kern-.05em{\sc i\kern-.025em b}\kern-.08em
    T\kern-.1667em\lower.7ex\hbox{E}\kern-.125emX}}

\title{MamFusion: Multi-Mamba with Temporal Fusion for Partially Relevant Video Retrieval\\
\thanks{This work is supported by National Key R\&D Program of China (2022ZD0119100), National Natural Science Foundation of China (Grant No: 62476238, 62202436), and the Natural Science Foundation of Zhejiang Province of China under Grant(No.LY24F020012, LHZSD24F020001).}
}

\author{
\IEEEauthorblockN{
Xinru Ying\textsuperscript{1,†}, 
Jiaqi Mo\textsuperscript{2,†}, 
Jingyang Lin\textsuperscript{1}, 
Canghong Jin\textsuperscript{1,4}, 
Fangfang Wang\textsuperscript{3}, 
Lina Wei\textsuperscript{$\ast$,1,4}
}
\thanks{†Equal contribution}
\thanks{$\ast$Corresponding author: \texttt{weiln@hzcu.edu.cn}}
\IEEEauthorblockA{
\textsuperscript{1}School of Computer and Computing Science, Hangzhou City University, Hangzhou, China \\
\textsuperscript{2}College of Letters \& Science, {University of Wisconsin\textendash Madison}, WI, USA \\
\textsuperscript{3}School of Information Science and Technology, Hangzhou Normal University, Hangzhou, China \\
\textsuperscript{4}Zhejiang Provincial Engineering Research Center for Real-Time SmartTech in Urban Security Governance \\
\texttt{Email:}\texttt{32101009@stu.hzcu.edu.cn, mo29@wisc.edu, 32201115@stu.hzcu.edu.cn}\\
\texttt{jinch@hzcu.edu.cn, wangff@hznu.edu.cn
}
}
}
\begin{document}
\maketitle

\begin{abstract}
Partially Relevant Video Retrieval (PRVR) is a challenging task in the domain of multimedia retrieval. It is designed to identify and retrieve untrimmed videos that are partially relevant to the provided query. In this work, we investigate long-sequence video content understanding to address information redundancy issues. Leveraging the outstanding long-term state space modeling capability and linear scalability of the Mamba module, we introduce a multi-Mamba module with temporal fusion framework (MamFusion) tailored for PRVR task. This framework effectively captures the state-relatedness in long-term video content and seamlessly integrates it into text-video relevance understanding, thereby enhancing the retrieval process. Specifically, we introduce Temporal T-to-V Fusion and Temporal V-to-T Fusion to explicitly model temporal relationships between text queries and video moments, improving contextual awareness and retrieval accuracy. Extensive experiments conducted on large-scale datasets demonstrate that MamFusion achieves state-of-the-art performance in retrieval effectiveness. Code is available at the link: \url{https://github.com/Vision-Multimodal-Lab-HZCU/MamFusion}.
\end{abstract}


\section{Introduction}
\label{sec:intro}

\begin{figure}[htbp] 
\setlength{\textfloatsep}{1pt} 
\setlength{\abovecaptionskip}{2pt} 
\setlength{\belowcaptionskip}{2pt} 
\centering
\includegraphics[width=\columnwidth]{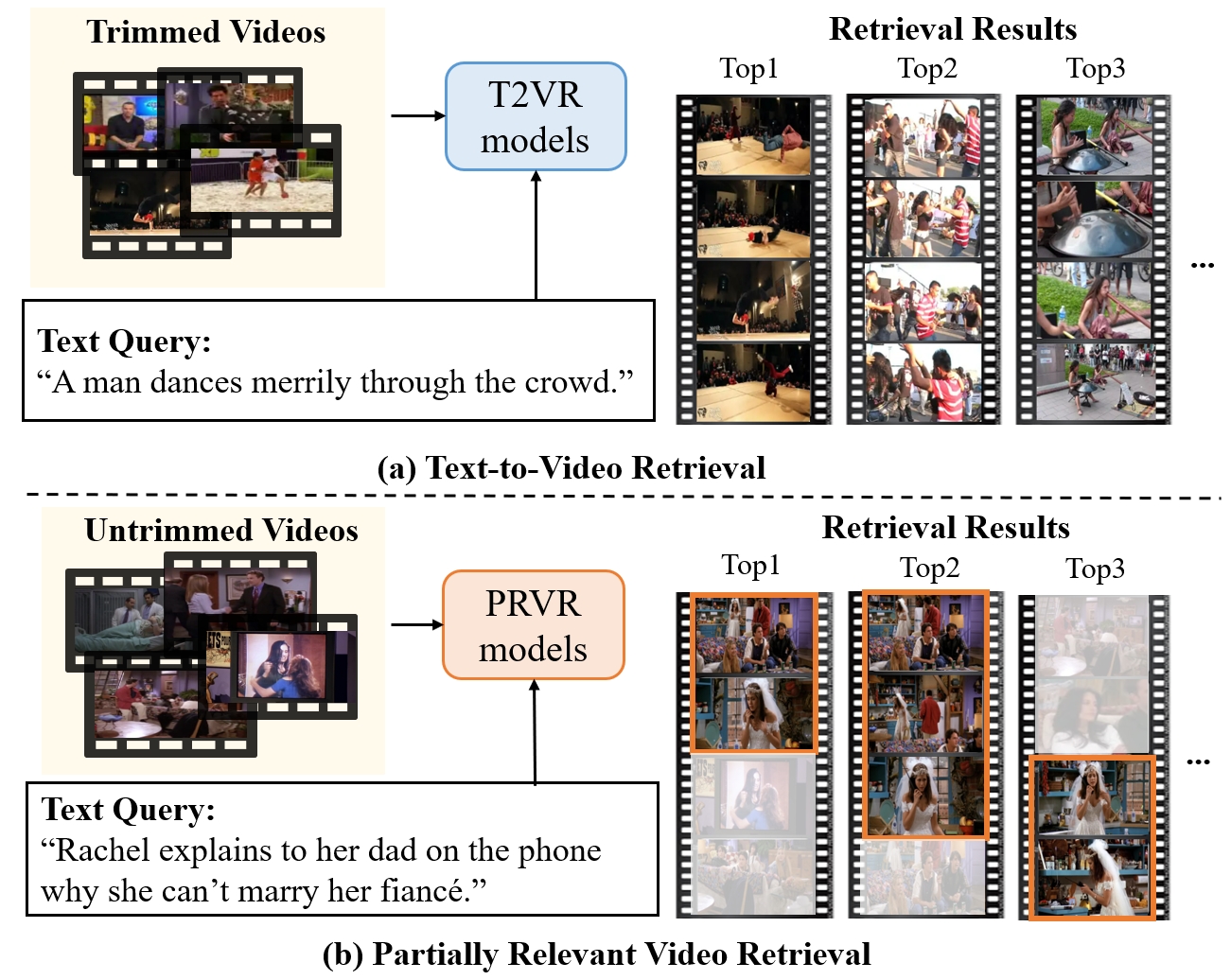} 
\caption{\textbf{An illustrative of conventional Text-to-video Retrieval and Partially Relevant Video Retrieval}: (a) The T2VR retrieves video files
from a trimmed video collection. (b) The PRVR retrieves video files from an unmodified long video collection and the retrieved video is partially associate with a text query.}
\label{fig:PRVR}
\vspace{-3.5mm}
\end{figure}

Text-to-video retrieval (T2VR) is a crucial task in video understanding and cross-modal information retrieval. 
Though many successful T2VR methods has been seen in recent research that excel in retrieving fully relevant short videos with natural language queries \cite{bain2021frozen, chen2020fine, dong2021dual, jin2021hierarchical, yang2021deconfounded, wang2022cross}, it remains challenging to conduct Partially Relevant Video Retrieval (PRVR)\cite{dong2022partially}, which deals with long untrimmed videos of only partial relevance to the query (see Fig.~\ref{fig:PRVR}). 
Unlike traditional T2VR task, PRVR aims to extract relevant video moments instead of entire videos, which is closer to real-world situations.

The emergence of large-scale video datasets and advancement in cross-modal retrieval have accelerated the development of both T2VR and PRVR fields. The existing text-video retrieval methods \cite{bain2021frozen, chen2020fine, jin2021hierarchical, ge2022bridging, wang2021t2vlad,fang2024prota} relies on video summaries and paired descriptions, using datasets such as MSVD \cite{chen2011collecting}, MSR-VTT \cite{xu2016msr}, and VATEX \cite{wang2019vatex}. These datasets typically assume short, trimmed videos with a direct alignment between video content and text descriptions, while in practical applications videos are often long and untrimmed, with only specific segments relevant to the query.

To address this, Video-Collection Moment Retrieval (VCMR) task \cite{paul2021text, wang2022siamese, zhang2020hierarchical} was introduced to retrieve semantically relevant moments from untrimmed video collections using textual queries. Most state-of-the-art VCMR methods employ a two-step process: using late fusion models to retrieve relevant videos, followed by early fusion models to locate specific moments \cite{paul2021text}. While these methods have shown effectiveness in moment-level retrieval, they still face challenges in PRVR, where precise video clips must be directly retrieved from large untrimmed video collections.

The original PRVR model uses multi-scale video representation and clip construction with Transformers for frame-level features \cite{dong2022partially}, resulting in many irrelevant clip embeddings, causing inefficiencies and storage overhead. GMMFormer \cite{wang2024GMMFormer} addresses this by applying Gaussian Mixture Model (GMM) constraints to compactly represent clip embeddings and reduce storage costs. DL-DKD \cite{dong2023dual} employs Dynamic Knowledge Distillation to enhance retrieval accuracy, particularly in data-scarce scenarios. PEAN \cite{jiang2023progressive} aligns text queries with local video content through three progressive modules to improve accuracy and efficiency. However, they still faces challenges in handling the redundancy and complexity of long, untrimmed video sequences.

To this end, we propose a Multi-Mamba module with Temporal Text-to-Video and Temporal Video-to-Text Fusion \cite{li2023catr} to incorporate the multi-modal information into our PRVR method. Specifically, the Multi-Mamba module dynamically adjusts model parameters to focus on relevant segments, and thus reducing redundancy. Besides, to handle the complex video-to-text interaction, we design the Temporal V-to-T Fusion module. This module replaces the traditional attention mechanism in the text representation with a temporal fusion mechanism, allowing the model to better understand the temporal context of the video moments in relation to the text query. In addition, to achieve comprehensive and balanced representation alignment, Temporal T-to-V Fusion is designed to capturing the temporal relationships from text queries to video moments. This module is introduced after the Mamba Block in the clip representation, allowing the model to reduces model redundancy and improves the computation economically. Our MamFusion method employs a top-down approach to extract effective information from long-term state spaces, thereby capturing potential video events. Simultaneously, it integrates the two modalities through cross-modal fusion guided by both video and text.  Extensive experiments on three public benchmark datasets were conducted to showcase the superior performance of our method.


In summary, our main contributions are as follows:

\begin{itemize}
    \item We propose a Multi-Mamba module which allows the model to dynamically focus on relevant segments and thus effectively solve the information redundancy problem of long-sequence video understanding; 
    \item We introduce two temporal fusion modules, Temporal T-to-V Fusion and Temporal V-to-T Fusion, which comprehensively models the complex interaction between video and text representations, bringing better contextual understanding; 
    \item Extensive experiments demonstrate that MamFusion significantly outperforms existing methods and achieves state-of-the-art performances in retrieval accuracy. 
\end{itemize}

\section{Our Method}

\begin{figure*}[htbp] 

\centering
\includegraphics[width=0.8\textwidth]{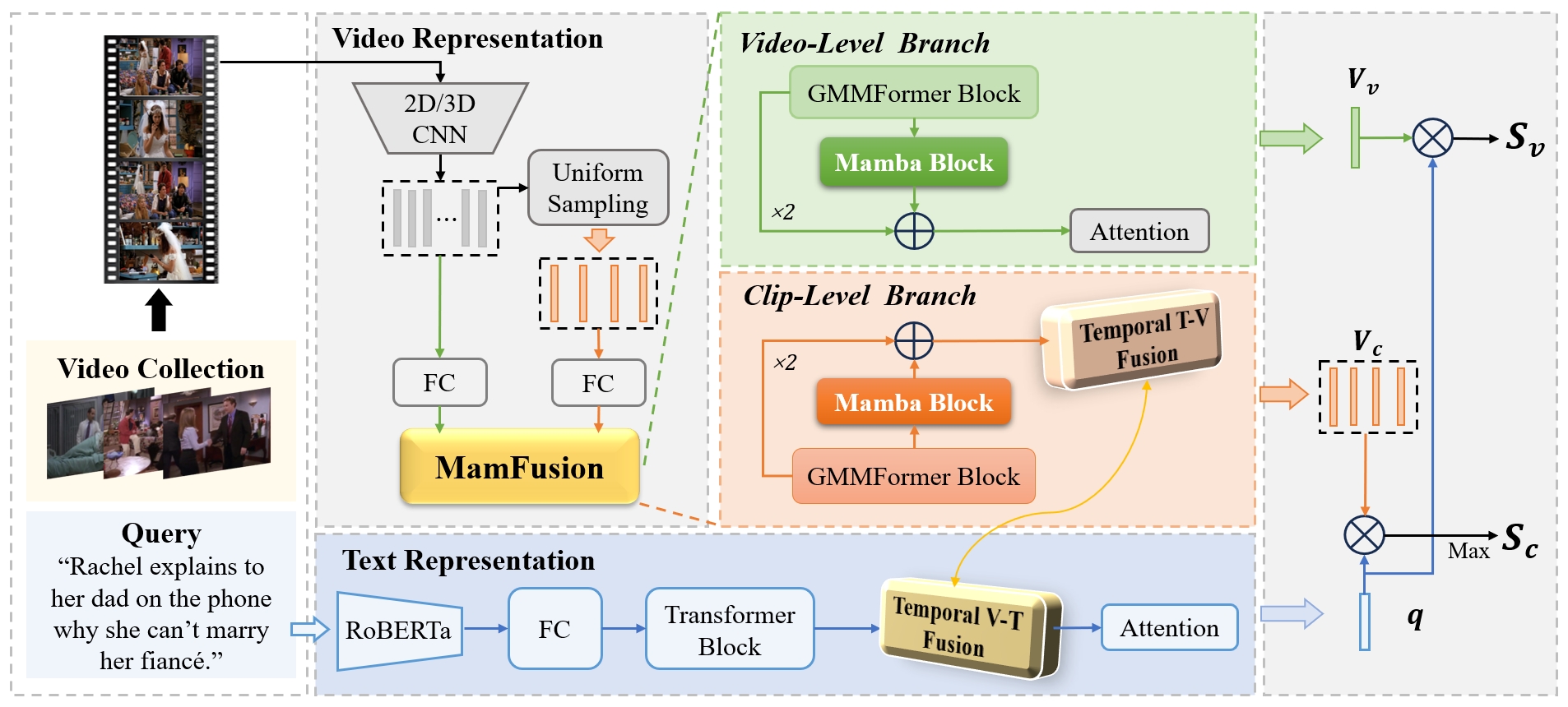} 
\caption{\textbf{An overview of our proposed MamFusion framework.}
Text features are extracted using RoBERTa, while video features are extracted using a 2D/3D CNN. The video representation is processed through clip-level and video-level branches. Temporal V-to-T Fusion and Temporal T-to-V Fusion enhance text-video interaction by capturing temporal context. The Mamba module dynamically focuses on relevant video moments, and cosine similarity is used for retrieval.}
\label{fig:model}
\vspace{-5mm}
\end{figure*}

In this section, we introduce our enhanced model for Partially Relevant Video Retrieval (PRVR) task, which integrates the Mamba module and Temporal V-to-T Fusion/Temporal T-to-V Fusion modules to improve text-video interaction, shown in Fig.~\ref{fig:model}. 

\subsection{Sentence Representation}

For sentence representation, we use the pre-trained RoBERTa model\cite{liu2019roberta} to extract word features from a given sentence. The sentence, composed of $N$ words, is first processed by the RoBERTa model to generate a sequence of word features. These features are then passed through a fully connected (FC) layer with a ReLU activation function to reduce their dimensionality. To incorporate positional information, we add a learned positional embedding to the mapped features. Subsequently, a standard Transformer layer is employed to obtain a sequence of $d$-dimensional contextualized word feature vectors $Q = \left\{ q_{i} \right\}_{i = 1}^{N} \in \mathbb{R}^{N \times d}$. The Transformer layer consists of a multi-head attention mechanism followed by a feed-forward network, both of which are connected with residual connections\cite{he2016deep} and layer normalization\cite{lei2016layer}. Finally, a sentence-level representation $q \in \mathbb{R}^{d}$ is derived by applying a simple attention mechanism on $Q$:
\begin{equation}
q = \sum_{i=1}^N \alpha_i^q \times q_i, \quad \alpha^q = \operatorname{Softmax}\left(w Q^T\right),
\end{equation}
where $\text{Softmax}$ denotes the softmax function, $w \in \mathbb{R}^{1 \times d}$ is a trainable vector, and $\alpha^{q} \in \mathbb{R}^{1 \times N}$ represents the attention weights.

\subsection{Multi-Scale Video Representation}

Given an untrimmed video with $M_{f}$ frames, we first extract frame features using a pre-trained 2D or 3D CNN. These features are then processed through two branches to obtain both clip and video embeddings. The clip embeddings aid in identifying relevant moments, while the video embeddings assess the overall text-video similarity.

\subsubsection{Clip-Level Branch}

Existing PRVR methods suffer from explicit clip modeling redundancy, which generates a large number of irrelevant clip embeddings and increases storage overhead. To overcome this, we design the clip-level branch to implicitly model clip representations. In this branch, we uniformly sample a fixed number of feature vectors by averaging over corresponding consecutive frame features. A fully connected layer with a ReLU activation is then used to reduce the dimensionality, resulting in clip features. Finally, we apply two GMMFormer blocks with learnable positional embeddings to these clip features to obtain clip embeddings $V_{c} = \left\{ c_{i} \right\}_{i = 1}^{M_{c}} \in \mathbb{R}^{M_{c} \times d}$, where $M_{c}$ is the number of sampled clips and $d$ is the feature dimension.

\subsubsection{Video-Level Branch}

To address the global frame interaction confusion in PRVR, we design the video-level branch to model the overall text-video similarity. In this branch, we first reduce the dimensionality using a fully connected layer with a ReLU activation. We then apply two GMMFormer layers with learnable positional embeddings to obtain contextualized features $V_{f} = \left\{ v_{i} \right\}_{i = 1}^{M_{f}} \in \mathbb{R}^{M_{f} \times d}$. Finally, we apply a simple attention mechanism on $V_{f}$ to derive video embeddings $V_{v} \in \mathbb{R}^{d}$:
\begin{equation}
V_v = \sum_{i=1}^{M_f} \alpha_i^f \times v_i, \quad \alpha^f = \operatorname{Softmax}\left(w V_f^T\right),
\end{equation}
where $w \in \mathbb{R}^{1 \times d}$ is a trainable vector and $\alpha^{f} \in \mathbb{R}^{1 \times M_{f}}$ represents the attention weights.

\subsubsection{GMMFormer Module}

To model temporal dependencies and capture multi-scale clip information, we introduce the GMMFormer module\cite{wang2024GMMFormer}. This module leverages Gaussian constraints to focus each frame on its adjacent frames, thereby enhancing the representation of the temporal context. Specifically‌, the GMMFormer module applies Gaussian attention to project these features into query, key, and value matrices which ensures that each frame focuses on its adjacent frames, enhancing temporal context representation.

The output of the Gaussian attention is then processed by a Feed-Forward Network (FFN) with residual connections and Layer Normalization. To handle varying moment lengths effectively, the module aggregates multiple Gaussian blocks with different variances, capturing multi-scale temporal information across clips. This aggregation helps maintain the temporal coherence of video sequences while encoding key information at different scales.

\subsection{Temporal Fusion Modules}

To address the challenge of effectively capturing the temporal relationships between text queries and video moments, inspired by \cite{li2023catr,xu2024multi}, we introduce two complementary modules: Temporal V-to-T Fusion and Temporal T-to-V Fusion. Specifically, the module first performs video-to-text fusion to integrate video features with text query features, followed by text-to-video fusion to further refine the alignment between text queries and video features. This sequential fusion process improves the model's ability to capture temporal relationships between video moments and text queries.

\subsubsection{Temporal V-to-T Fusion}

The Temporal T-to-V Fusion module operates by fusing the text representation with the video representation at each time step, ensuring that the text representation is dynamically updated based on the video content. This fusion process is performed using a multi-head attention mechanism, where the text representation serves as the query and the video representation serves as the key and value. The resulting fused representation captures the temporal dependencies between the text query and the video moments, leading to more accurate and context-aware retrieval results. After generating the context word feature vector $Q = \left\{ q_{i} \right\}_{i = 1}^{N} \in \mathbb{R}^{N \times d}$ through the Transformer layer, the generated text representation $Q$ is fused with the video representation $V_{f_{m}}$ to generate a new text representation $Q^{'}$. The specific formula is as follows:
\begin{equation}
\scalebox{0.85}{$\displaystyle
\mathrm{Q}^{\prime}=\operatorname{TVT}\left(Q, V_{f_t}\right) = \operatorname{Softmax}\left(\frac{Q W^Q (V_{f_t} W^K)^T}{\sqrt{d_{\text{head}}}}\right) V_{f_t} W^V,
$}
\end{equation}
where TVT(·) denotes the Temporal Video-to-Text Fusion function that enables video features to enhance text representations, $Q$ denotes the text representation, $V_{f_{t}}$ represents the video features processed by the Mamba module, $W^Q$, $W^K$, $W^V$ are learnable parameter matrices, and $d_\text{head}$ is the attention head dimension.

\subsubsection{Temporal T-to-V Fusion}

The Temporal V-to-T Fusion module operates by fusing the video representation with the text representation at each time step, ensuring that the video representation is dynamically updated based on the text query. This fusion process is performed using a multi-head attention mechanism\cite{vaswani2017attention}, where the video representation serves as the query and the text representation serves as the key and value. The resulting fused representation captures the temporal dependencies between the video moments and the text query, leading to more accurate and context-aware retrieval results. The generated video representation $V_{f_m}$ is fused with the text representation $q$ to generate a new video representation $V_{f_{t}}$. The specific formula is as follows:
\begin{equation}
\scalebox{0.85}{$\displaystyle 
V_{f_t}=\operatorname{TTV}\left(V_{f_m}, q\right) = \operatorname{Softmax}\left(\frac{V_{f_m} W^Q (q W^K)^T}{\sqrt{d_{\text{head}}}}\right) q W^V,
$}
\end{equation}

where TTV(·) represents the Temporal Text-to-Video Fusion function that incorporates textual information into video representations, $V_{f_m}$ is the video representation, $q$ is the text representation.

\subsection{Multi-Mamba Module}

To address the challenge of dynamic processing in PRVR, where the model needs to handle long video sequences and complex text queries, we introduce the Multi-Mamba Module to manage large-scale, untrimmed video retrieval.

The Mamba module dynamically adjusts parameters to focus on relevant moments and ignore irrelevant ones, enabling effective handling of long video sequences and complex queries in PRVR tasks. The video feature \(V_{f}\) is processed through the GMMFormer module to produce \(V_{f_{g}}\), which is then passed into the Mamba module to generate the final video representation \(V_{f_{m}}\):
\begin{equation}
V_{f_m}=\operatorname{Mamba}\left(V_{f_g}\right) .
\end{equation}

This integration allows the model to adapt to the video's temporal structure, effectively processing long sequences and improving performance in PRVR tasks.

\section{Experiments}

\subsection{Experimental Settings}

\subsubsection{Datasets}

To validate the effectiveness of our proposed PRVR (Partial Relevant Video Retrieval) model, we utilized three large-scale video datasets: ActivityNet Captions\cite{krishna2017dense}, Charades-STA\cite{gao2017tall}, and TVR\cite{lei2020tvr}. These datasets are designed to evaluate the ability of models to retrieve videos based on queries that are relevant to specific segments of the videos. Notably, since our task focuses on video retrieval rather than moment localization, the moment annotations provided by these datasets were not used in our experiments.

\subsubsection{Evaluation Metrics}

Following \cite{dong2022partially}, we use R@K (K = 1, 5, 10, 100) as the primary evaluation metric. R@K measures the percentage of queries for which the desired video appears in the top-K ranked results. Additionally, we report SumR, which is the sum of all recall rates (R@1, R@5, R@10, R@100). A higher SumR indicates better overall performance.

\subsubsection{Implementation Details}

For the video representations, we utilized I3D features for the ActivityNet Captions and Charades-STA datasets, as provided by \cite{zhang2020hierarchical} and \cite{mun2020local}, respectively. For the TVR dataset, we employed 3,072-dimensional visual features, which were obtained by concatenating frame-level ResNet152 features \cite{he2016deep} and segment-level I3D features \cite{carreira2017quo}, as provided by \cite{lei2020tvr}.

For sentence representations, we used 1,024-dimensional RoBERTa features for ActivityNet Captions and Charades-STA, extracted by \cite{dong2022partially}. For TVR, we utilized 768-dimensional RoBERTa features, also provided by \cite{lei2020tvr}.

For the Gaussian blocks, which include low, medium, high, and infinite variance settings, we set the variances to 0.5, 1.0, 5.0, and $\infty$ respectively. Additionally, for the Mamba module, we configured the state expansion factor $d_{\text{state}}$, local convolution width $d_{\text{conv}}$, and block expansion factor $\text{expand}$. Specifically, we set $d_{\text{state}}$ to 16, $d_{\text{conv}}$ to 4, and $\text{expand}$ to 2. These parameters were optimized to enhance the model's ability to capture temporal dependencies and improve the representation of video segments.

The Mamba module was integrated into both the clip-level and video-level branches of our model. In the clip-level branch, the Mamba module processed the sampled clip features, while in the video-level branch, it was applied to the entire sequence of frame features. This dual integration allowed the model to balance local and global temporal information, leading to more robust and accurate video retrieval.

\subsection{Main Results}

We compared our proposed MamFusion model with several state-of-the-art PRVR methods, as well as with T2VR and VCMR models. Specifically, we evaluated the following T2VR models: RIVRL\cite{dong2022reading}, CLIP4Clip\cite{luo2022clip4clip}, and Cap4Video\cite{wu2023cap4video}, and the following VCMR models: ReLoCLNet\cite{zhang2021video}, XML\cite{lei2020tvr}, and CONQUER\cite{hou2021conquer}.

\begin{figure}[htbp] 
\centering
\vspace{-1.5mm}
\begin{minipage}{0.49\columnwidth}
    \centering
    \includegraphics[width=\linewidth]{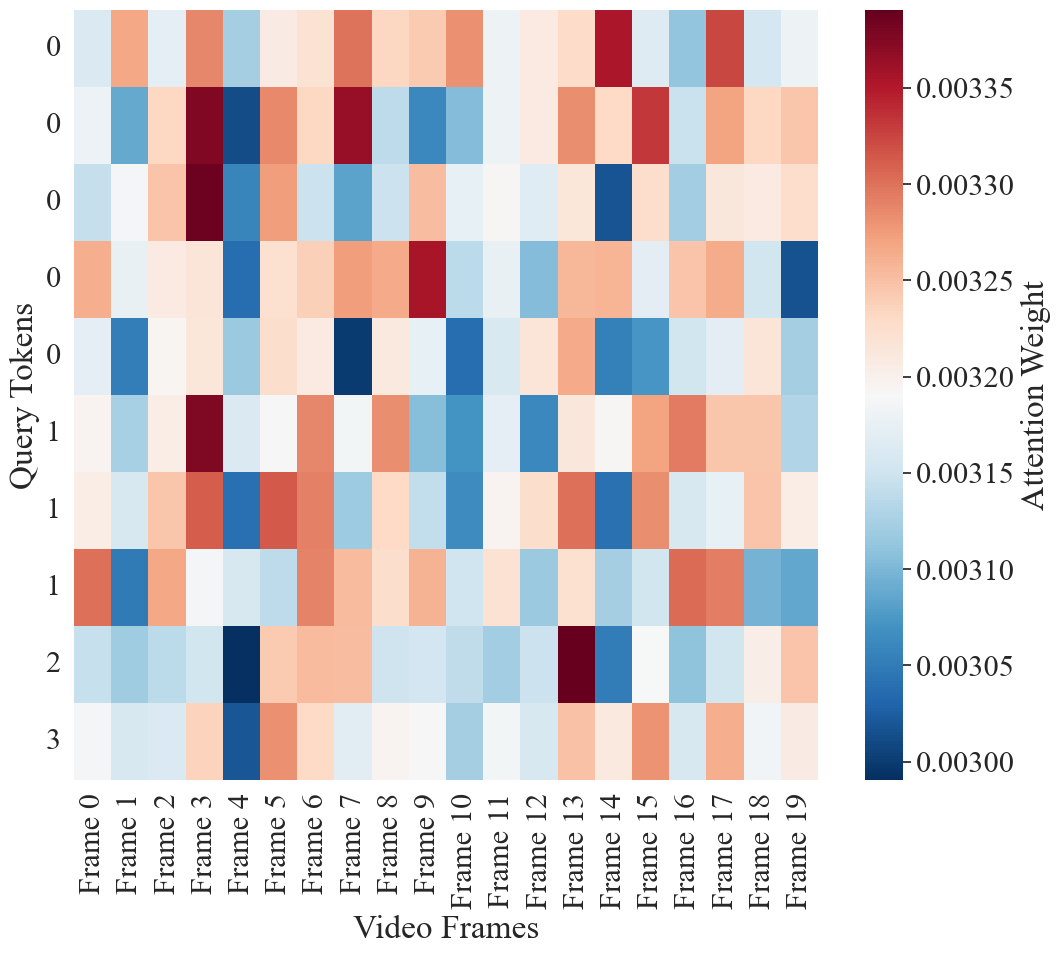} 
    (a) Video-to-Text Mapping 
\end{minipage}
\hfill
\begin{minipage}{0.49\columnwidth}
    \centering
    \includegraphics[width=\linewidth]{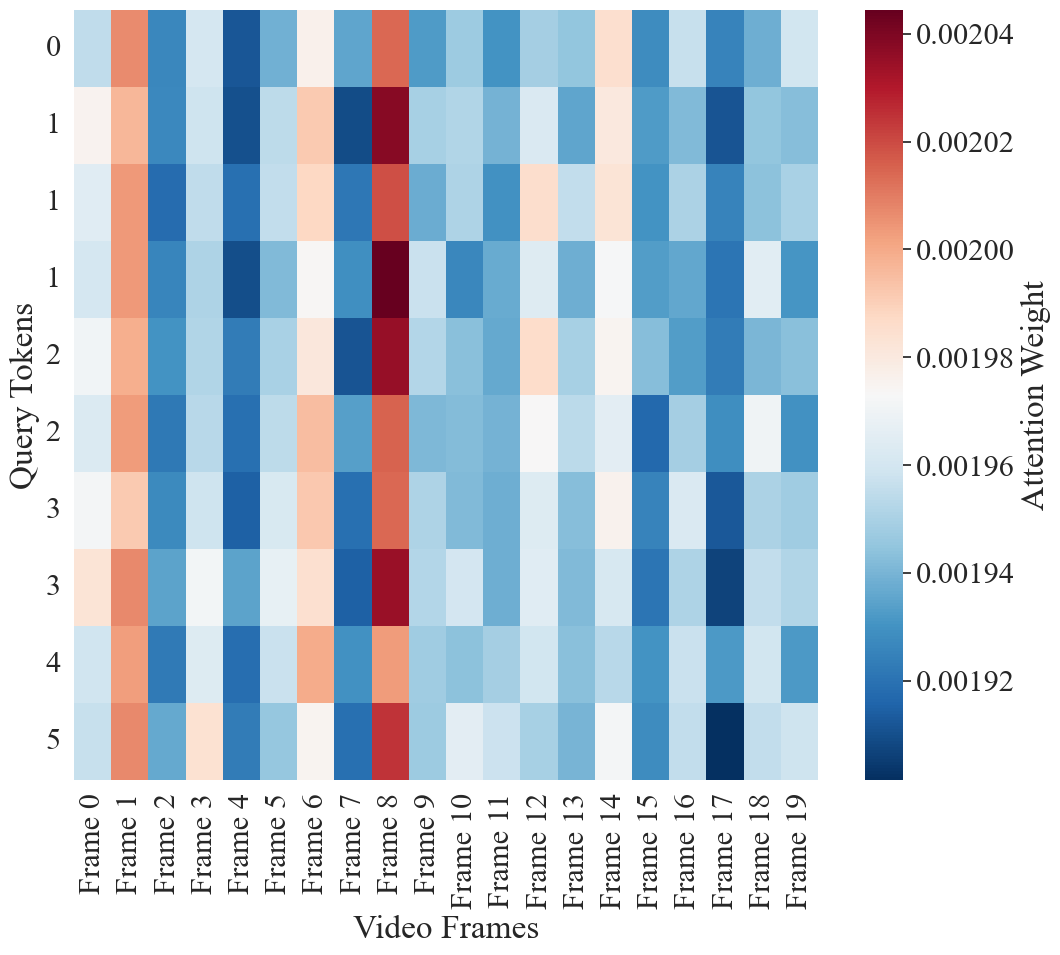} 
    (b) Text-to-Video Mapping 
\end{minipage}
\caption{Attention Heatmaps for Video-to-Text and Text-to-Video Mappings. The heatmaps illustrate the attention weights between video frames and text query tokens (a) and between text query tokens and video frames (b). Darker regions in the heatmaps indicate higher relevance or stronger temporal relationships, demonstrating the model's ability to focus on relevant segments and improve retrieval accuracy.}
\label{fig:heatmaps}
\vspace{-3.8mm}
\end{figure}

Table~\ref{tab:three datasets} summarize the performance of various models on the ActivityNet Captions, Charades-STA, and TVR datasets, respectively. Our MamFusion model achieved state-of-the-art performance across all datasets, outperforming both T2VR and VCMR models. To further analyze the alignment between text queries and video frames, we visualized the attention weights using heatmaps. As shown in Figure \ref{fig:heatmaps}, these visualizations reveal distinct patterns: Video-to-Text mapping (a) shows dynamically changing attention across time, reflecting how TVT integrates evolving video content into text representations; while Text-to-Video mapping (b) exhibits concentrated attention on specific frames, demonstrating TTV's precise localization capability. These complementary patterns validate our bidirectional fusion mechanism's effectiveness in capturing temporal relationships between modalities. 


\begin{table*}[htbp]
\caption{Performance of various models on ActivityNet Captions, Charades-STA, and TVR datasets. Note that the experimental results for these models are localized across the three datasets.}
\setlength{\textfloatsep}{1pt} 
\setlength{\abovecaptionskip}{0.5pt} 
\setlength{\belowcaptionskip}{0.5pt} 
\begin{center}
\setlength{\tabcolsep}{4pt} 
\begin{tabular}{|l|c|c|c|c|c|c|c|c|c|c|c|c|c|c|c|}
\hline
\textbf{Model} & \multicolumn{5}{c|}{\textbf{ActivityNet Captions}} & \multicolumn{5}{c|}{\textbf{Charades-STA}} & \multicolumn{5}{c|}{\textbf{TVR}} \\ \hline
 & R@1 & R@5 & R@10 & R@100 & SumR & R@1 & R@5 & R@10 & R@100 & SumR & R@1 & R@5 & R@10 & R@100 & SumR \\ \hline
\multicolumn{16}{|l|}{\textit{T2VR models:}} \\ \hline
RIVRL & 5.2 & 18.0 & 28.2 & 66.4 & 117.8 & 1.6 & 5.6 & 9.4 & 37.7 & 54.3 & 9.4 & 23.4 & 32.2 & 70.6 & 135.6 \\
CLIP4Clip & 5.9 & 19.3 & 30.4 & 71.6 & 127.3 & 1.8 & 6.5 & 10.9 & 44.2 & 63.4 & 9.9 & 24.3 & 34.3 & 72.5 & 141.0 \\
Cap4Video & 6.3 & 20.4 & 30.9 & 72.6 & 130.2 & 1.9 & 6.7 & 11.3 & 45.0 & 65.0 & 10.3 & 26.4 & 36.8 & 74.0 & 147.5 \\ \hline
\multicolumn{16}{|l|}{\textit{VCMR models:}} \\ \hline
ReLoCLNet & 5.7 & 18.9 & 30.0 & 72.0 & 126.6 & 1.2 & 5.4 & 10.0 & 45.6 & 62.3 & 10.7 & 28.1 & 38.1 & 80.3 & 157.1 \\
XML & 5.3 & 19.4 & 30.6 & 73.1 & 128.4 & 1.6 & 6.0 & 10.1 & 46.9 & 64.6 & 10.0 & 26.5 & 37.3 & 81.3 & 155.1 \\
CONQUER & 6.5 & 20.4 & 31.8 & 74.3 & 133.1 & 1.8 & 6.3 & 10.3 & 47.5 & 66.0 & 11.0 & 28.9 & 39.6 & 81.3 & 160.8 \\ \hline
\multicolumn{16}{|l|}{\textit{PRVR models:}} \\ \hline
MS-SL & 7.1 & 22.5 & 34.7 & 75.8 & 140.1 & 1.8 & 7.1 & 11.8 & 47.7 & 68.4 & 13.5 & 32.1 & 43.4 & 83.4 & 172.4 \\
JSG & 6.8 & 22.7 & 34.8 & 76.1 & 140.5 & \textbf{2.4} & 7.7 & 12.8 & 49.8 & 72.7 & - & - & - & - & - \\
GMMFormer & \textbf{8.3} & 24.9 & 36.7 & 76.1 & 146.0 & 2.1 & 7.8 & 12.5 & 50.6 & 72.9 & 13.9 & 33.3 & 44.5 & \textbf{84.9} & 176.6 \\
\textbf{MamFusion} & 8.0 & \textbf{25.4} & \textbf{37.2} & \textbf{76.8} & \textbf{147.4} & 2.0 & \textbf{8.8} & \textbf{14.2} & \textbf{51.5} & \textbf{76.5} & \textbf{14.2} & \textbf{33.9} & \textbf{44.9} & 84.5 & \textbf{177.5} \\ \hline
\end{tabular}
\label{tab:three datasets}
\end{center}
\vspace{-5mm}
\end{table*}

\subsection{Ablation Studies}

To further validate the effectiveness of the proposed MamFusion model, we conducted a series of ablation studies to evaluate the contributions of its key components: the Multi-Mamba Module, the Temporal V-to-T Fusion, and the Temporal T-to-V Fusion. These studies aim to demonstrate the importance of each module in improving retrieval performance.

\subsubsection{Multi-Mamba Module}

The Multi-Mamba block is a core component of MamFusion, designed to enhance the model's dynamic processing capabilities by leveraging selective state space models. To assess its impact, we conducted ablation experiments by replacing the Multi-Mamba block with the baseline model, GMMFormer, and comparing the performance with the full MamFusion model. The results of the ablation study are presented in Table~\ref{tab:ablation_combined}. As shown, replacing the Multi-Mamba with the GMMFormer baseline significantly degraded the retrieval performance, indicating that the dynamic processing capabilities provided by the Multi-Mamba Module are crucial for handling long video sequences and complex queries.

\subsubsection{Temporal Fusion Modules}

The Temporal T-to-V Fusion and Temporal V-to-T Fusion modules are designed to enhance the interaction between text and video representations by explicitly modeling the temporal context. To evaluate their contributions, we conducted ablation experiments by removing each module individually and comparing the results with the full model. The results of the ablation study are presented in Table~\ref{tab:ablation_combined}. As shown, removing either the Temporal V-to-T Fusion or the Temporal T-to-V Fusion module resulted in a significant drop in retrieval performance, indicating that both modules play a crucial role in capturing the temporal relationships between text queries and video moments.

\begin{table}[htbp]
\vspace{-1.5mm}
\caption{Ablation studies of MamFusion on Charades-STA. V2T means Temporal V-to-T Fusion and T2V means Temporal T-to-V Fusion.}
\vspace{-5mm}
\begin{center}
\resizebox{\columnwidth}{!}{ 
\begin{tabular}{|c|c|c|c|c|c|}
\hline
\textbf{Model Variant} & \textbf{R@1} & \textbf{R@2} & \textbf{R@10} & \textbf{R@100} & \textbf{SumR} \\
\hline
w/o Multi-Mamba & 1.6 & 6.9 & 11.2 & 49.8 & 69.5 \\
\hline
w/o V2T & \textbf{2.3} & 7.9 & 13.1 & 51.3 & 74.7 \\
\hline
w/o T2V & 2.0 & 7.9 & 12.4 & 50.2 & 72.6 \\
\hline
w/o Both Temporal Fusions & 2.1 & 7.5 & 12.7 & 49.7 & 72.0 \\
\hline
MamFusion(Full) & 2.0 & \textbf{8.8} & \textbf{14.2} & \textbf{51.5} & \textbf{76.5} \\
\hline
\end{tabular}
}
\label{tab:ablation_combined}
\end{center}
\vspace{-3mm}
\end{table}

The ablation studies demonstrate that each component of MamFusion plays a crucial role in improving retrieval performance. The Multi-Mamba block enhances dynamic processing, while the Temporal V-to-T Fusion and Temporal T-to-V Fusion modules improve text-video interaction by capturing temporal context. Together, these modules enable MamFusion to achieve state-of-the-art performance in partially relevant video retrieval tasks.

\subsection{Inference Convergence Speed}

To further evaluate MamFusion's performance during training, we compared the convergence speed of MamFusion and GMMFormer. Specifically, we measured the average loss convergence during the training process.

\begin{figure}[htbp] 
\vspace{-6mm}
\centering
\includegraphics[width=0.66\columnwidth]{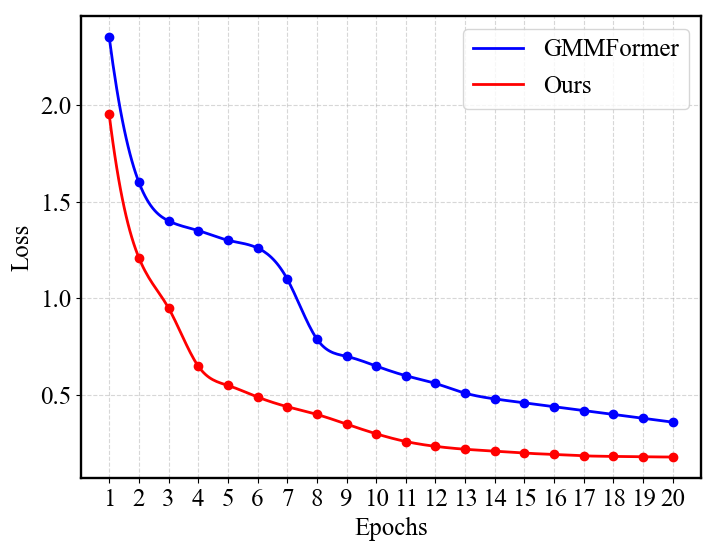} 
\caption{A comparison of the average loss convergence during the training process for MamFusion and GMMFormer. As shown, MamFusion converges faster and more smoothly, reaching a lower loss value in fewer epochs, indicating its superior retrieval performance and stability during training.}
\label{fig:shoulian}
\vspace{-3.6mm}
\end{figure}

As shown in Figure \ref{fig:shoulian}, MamFusion demonstrates a significantly faster and smoother convergence compared to GMMFormer. The average loss of MamFusion decreases at a noticeably faster rate, reaching a lower value in fewer epochs. This indicates that MamFusion not only converges more quickly but also achieves a more stable reduction in loss during training. These improvements highlight MamFusion’s superior ability to adapt to the task, making it suitable for practical applications.

\section{Conclusion}

This paper introduces MamFusion, a novel model for Partially Relevant Video Retrieval (PRVR), which integrates the Multi-Mamba module with Temporal V-to-T Fusion and Temporal T-to-V Fusion, built on top of GMMFormer. The Multi-Mamba module enhances dynamic processing, enabling better handling of long video sequences and complex queries. Experimental results demonstrate that MamFusion outperforms existing methods in retrieval accuracy. The unique capabilities of the Mamba module in dynamic processing and compact representation make it a key innovation for improving PRVR tasks.


\vspace{-2mm}
\bibliographystyle{IEEEbib}
\bibliography{icme2025references}

\end{document}